# On the Number of Experiments Sufficient and in the Worst Case Necessary to Identify All Causal Relations Among $N$ Variables

**Frederick Eberhardt, Clark Glymour and Richard Scheines**
Department of Philosophy
Carnegie Mellon University
Pittsburgh, PA 15213

## Abstract

We show that if any number of variables are allowed to be simultaneously and independently randomized in any one experiment, $\log_2(N) + 1$ experiments are sufficient and in the worst case necessary to determine the causal relations among $N \geq 2$ variables when no latent variables, no sample selection bias and no feedback cycles are present. For all $K$, $0 < K < \frac{1}{2}N$ we provide an upper bound on the number experiments required to determine causal structure when each experiment simultaneously randomizes $K$ variables. For large $N$, these bounds are significantly lower than the $N-1$ bound required when each experiment randomizes at most one variable. For $k_{max} < \frac{N}{2}$, we show that $(\frac{N}{k_{max}}-1)+\frac{N}{2k_{max}}\log_2(k_{max})$ experiments are sufficient and in the worst case necessary. We offer a conjecture as to the minimal number of experiments that are in the worst case sufficient to identify all causal relations among $N$ observed variables that are a subset of the vertices of a DAG.

## 1 THE PROBLEM

Some scientific problems concern large numbers of variables that can be individually or collectively manipulated experimentally. The idea of randomization of a particular variable goes back to Fisher who in the 1930s suggested that determining the value of the treatment variable by sampling from a known probability distribution independent of any of the other variables, removes confounding by eliminating the influence of all other causes on the treatment variable. In the case described here we are considering a large number of variables and we are trying to discover the causal relations among these variables. There is no uniquely defined (set of) treatment variables. Consequently we are faced with an optimization problem over a sequence of experiments: How do we choose our interventions to minimize the number of experiments required to discover all the causal relations? There has been a vast amount of work on the optimization of experiments in the literature on experimental design. However, it has focused on the optimization of the assignment of values to the (set of) treatment variables in order to most efficiently discover the causal structure. In our case we consider the choice of the (set of) treatment variables as part of the experiment and hence as part of the optimization over the sequence of experiments.

For example, in studies of gene regulation, the expression of one or more genes can be simultaneously suppressed. Although gene suppression is not randomization of expression, such possibilities suggest questions of independent theoretical interest: If at most $K$ of $N > K$ variables can be simultaneously and independently randomized in a single experiment, what is the minimal number of experiments sufficient, and in the worst case necessary, to determine the causal relations among the $N$ variables? What is the optimal value of $K$? We assume there are no unmeasured common causes of the $N$ variables, that the system is free of feedback, and that the independence relations true of the population are available in each experiment.

Murphy (2001) and, independently, Tong and Koller (2001), place a related question in the framework of Causal Bayes Net representations of causal dependencies (Spirtes, et al., 2000), and ask, given a probability distribution over directed acyclic graphs (DAGs), which intervention on a single variable would be expected to be most informative. They assume, as will we, that there are no latent variables, no sample selection bias, no feedback, and that the conditional independence relations in the data perfectly model the d-separation relations (Pearl, 1988) of a DAG whose directed edges $X \to Y$, represent the proposition that for some values of all other $N-2$ variables, $Y$ covaries with $X$ when $X$ is randomized.

An ideal intervention, as by randomization, on a variable $V$ in a causal system removes the influence of other variables in the system on $V$. The intervention forces a distribution on $V$, and thereby changes the joint distribution of all variables in the system that depend directly or indirectly on $V$ but does not change the conditional distribution of other variables given values of $V$. After the randomization, the associations of the remaining variables with $V$ provide information about which variables $V$ influences, but the intervention hides information about which variables influence $V$.

Under the same assumptions we make, various pointwise consistent algorithms (Spirtes, et al., 2000; Meek, 1996; Chickering, 2002) are known for obtaining the Markov equivalence class of a DAG from conditional independence relations among a set of passively observed variables. Informally, our problem is to determine how best to combine the information from passive observations with the information from interventions.

More formally, our problem is as follows. Let $\mathbf{D}(N)$ be the set of all DAGs on $N$ vertices. Let a $K$-intervention on vertices $V_1, \ldots, V_k$, in a graph $G$ determine the subgraph $G \setminus \{V_1, \ldots, V_k\}$ in which all edges directed into $V_1, \ldots, V_k$ are removed. Then we say that an experiment consists in a choice of $K$ and a $K$-intervention. Let $\mathbf{O}$ be an oracle that returns for any $G$ and for any $K$-intervention experiment on $V_1, \ldots, V_k$ (including the null intervention experiment in which all variables in $V$ are passively observed), the independence relations implied by $G \setminus \{V_1, \ldots, V_k\}$. For all $N \geq 2$, what is the minimal number of experiments that suffice to uniquely identify any $G$ in $\mathbf{D}(N)$?

## 2  THE $K = 1, N = 3$ CASE

Our general strategy can be illustrated with the simplest case, $K = 1, N = 3$. There are 25 possible DAGs on three vertices, $V_1, V_2$ and $V_3$. There are two ways in which the direction of particular edge between $V_i$ and $V_j$ can be discovered. The first is a randomization of $V_i$. If $V_i$ and $V_j$ covary for all possible conditioning sets, then $V_i$ is a direct cause of $V_j$. The other case occurs when there is a third variable, $V_k$ such that $V_j$ is a common effect of $V_i$ and $V_k$ and $V_i$ and $V_k$ are not adjacent in the true graph. In this case we say that $V_j$ is an unshielded collider. Unshielded colliders are easily discovered since conditioning on the unshielded collider makes two previously independent variables dependent. This provides a strategy to direct edges into an unshielded collider.

Let $G$ (in Figure 1) be the true graph. Arbitrarily, begin with an experiment in which $V_1$ is randomized.

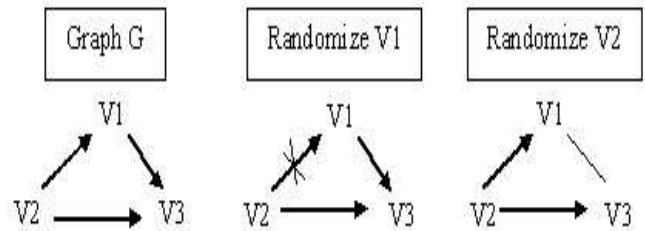

Figure 1: The true graph $G$ shown on the left, the equivalence class of graphs after a randomization of $V_1$ shown in the middle and the equivalence class of graphs after a randomization of $V_2$ on the right.

The resulting graph, $G \setminus \{V_1\}$, implies only one independence: $V_1 \perp V_2$. The Markov equivalence class for this independence *under passive observation* has only one member: $V_1 \rightarrow V_3 \leftarrow V_2$. In the case when $V_1$ is randomized, however, the equivalence class (middle of Figure 1) for this independence is more complicated. First, $V_1 \rightarrow V_3$, second, $V_2 \rightarrow V_3$ since $V_3$ is an unshielded collider in the post manipulation graph, and third, either $V_2 \rightarrow V_1$ or $V_2$ and $V_1$ are not adjacent (indicated by $V_2 \rightarrow V_1$ with an "X" through it). In general, in an experiment in which $V_1$ is randomized, $V_1$ will be associated with $V_j$ conditional on any subset iff $V_1 \rightarrow V_j$ in $G$, and $V_1$ will be independent of $V_j$ iff either $V_j \rightarrow V_1$ in $G$ or $V_1$ and $V_j$ are not adjacent in $G$. Unless we have unshielded colliders it will not in general be possible to direct edges that do not involve at least one variable that is intervened upon.

Next, arbitrarily, consider a different experiment in which $V_2$ is randomized. In this case the resulting graph $G \setminus \{V_2\}$ entails no independencies, and the equivalence class under the $V_2$ intervention is shown on the right of Figure 1. Notice, however, that we may now combine the results from both experiments. For the pair $V_2 - V_1$, the $V_1$ randomization experiment told us only that either $V_2 \rightarrow V_1$ or that $V_1$ and $V_2$ are not adjacent, but the $V_2$ randomization experiment determined that $V_2 \rightarrow V_1$. Similarly for the other two pairs, one of the experiments determined the nature of the connection between them uniquely. An effective sequential strategy of experiments in fact leverages the results of previous experiments to maximally reduce the size of the intersection of all the equivalence classes found so far. In this case, $N - 1$ experiments sufficed. The general result for single variable interventions follows from a more general theorem we will prove later.

**Proposition**: For $K = 1, N > 2$, $N - 1$ experiments

suffice to identify any DAG on $N$ vertices. Moreover, this is a best worst case lower bound that adaptive strategies cannot improve (Eberhardt, et al., 2004).

## 3 THE MINIMAL NUMBER OF EXPERIMENTS FOR ARBITRARY DAGs

We now consider the problem when we are free to randomize simultaneously as many variables as we please. The set of independence relations entailed by $G \setminus \{V_1, \ldots, V_k\}$ determines all of the adjacencies in $G$ among the variables in $V \setminus \{V_1, \ldots, V_k\}$ (Spirtes, et al., 2000) and all of the adjacencies and directions of edges from members of $\{V_1, \ldots, V_k\}$ to members of $V \setminus \{V_1, \ldots, V_k\}$ (Eberhardt, et al., 2004). In many cases, the Markov equivalence class of $G \setminus \{V_1, \ldots, V_k\}$ will also specify directions for some edges, but in the worst case, when $G$ is a complete graph, the Markov equivalence classes contain no such information, and so we assume no such information in seeking a worst case bound.

We say an experiment is a *directional test* for $V_j, V_k$, if exactly one of $V_j, V_k$ is randomized, an *adjacency test* for $V_j, V_k$ if neither is randomized, and a *zero information test* for $V_j, V_k$ if both are randomized. Two directional tests for $V_j, V_k$, are opposing if $V_j$ but not $V_k$ is randomized in one test, and $V_k$ but not $V_j$ is randomized in the other test. In order to determine all the relationships in a causal DAG of $N$ variables, each pair of variables $V_i$ and $V_j$ needs to be tested twice: either by two opposing directional experiments or by one directional and one adjacency experiment. Naively then, the problem of identifying the underlying causal structure is a matter of running two tests on each pair of variables.

**Lemma 3.1** *Let $G = (\mathbf{V}; \mathbf{E})$ be a graph on $N$ vertices and let $X$ be an experiment on $G$ consisting of a simultaneous intervention on $K \leq N$ variables. Let $\mathbf{I} \subset \mathbf{V}$ be the set of randomized variables, i.e. $|\mathbf{I}| = K$, and $\mathbf{U} = \mathbf{V} \setminus \mathbf{I}$. Then*

1. *$X$ is a directional test for $K(N-K)$ pairs of variables, namely all pairs $V_i, V_j$ where $V_i \in \mathbf{I}$ and $V_j \in \mathbf{U}$.*

2. *$X$ is an adjacency test for $\binom{(N-K)}{2}$ pairs of variables, namely all pairs $V_i, V_j \in \mathbf{U}$*

3. *$X$ is a zero-information test for $\binom{K}{2}$ pairs of variables, namely all pairs $V_i, V_j \in \mathbf{I}$.*

*Note that the number of pairs for which $X$ is a directional test is maximized at $k = \frac{N}{2}$.*

**Lemma 3.2** $\log_2(N)$ *experiments are sufficient to subject all pairs in a causal graph among $N$ variables to a directional test.*

**Proof:** Suppose that $N = 2^m$ for some positive integer $m$. Let $m = 1$, i.e. $N = 2$. Clearly one experiment, an intervention on one of the two variables, will subject this pair of variables and hence all the pairs in this graph to a directional test. Now suppose that the theorem holds for all $m \leq r$. Then let $m = r + 1$, i.e. $N = 2^{r+1}$. Let the first experiment $E_1$ consist of an intervention on $K = \frac{N}{2} = 2^r$ variables. It follows from Lemma 3.1, that $E_1$ is a directional experiment for the $\frac{N^2}{4}$ pairs of variables with one variable in $\mathbf{I}$ and the other in $\mathbf{U}$. Now, note that $|\mathbf{I}| = |\mathbf{U}| = K = \frac{N}{2} = \frac{2^{r+1}}{2} = 2^{m-1}$. By the induction hypothesis we know that for $N' = 2^{m-1}$, $\log_2(N')$ experiments are sufficient to subject all pairs in a causal graph among $N'$ variables to a directional test. Hence, within $r$ experiments, all pairs of variables in $\mathbf{I}$ and $\mathbf{U}$ have been subject to a directional test. Consequently, within $r + 1 = \log_2(N)$ experiments, all pairs of variables in $G$ have been subject to a directional experiment.

The bound also applies to the case where $N \neq 2^m$. The proof is more complex, but the intuition is the same: In order to accommodate all the directional tests, the aim is to split the original set of variables into equal sized subproblems that can be solved concurrently. That is, whenever the number N of variables in a problem is odd, the next intervention should occur on $\frac{N-1}{2}$ variables, resulting in subproblems of size $\frac{N-1}{2}$ and $\frac{N+1}{2}$. As a result it should be obvious that the number of recursive splits to accommodate all the directional tests for any $N$ is the same as for the closest $N^*$, where $N^* = 2^m$ and $m$ is the smallest integer such that $N \leq N^*$. □

**Theorem 3.3** $\log_2(N) + 1$ *experiments are sufficient to determine the causal graph among $N$ variables.*

**Proof:** From Lemma 3.2 it follows that $\log_2(N)$ experiments are sufficient to subject all pairs of variables in an $N$-variable graph to a directional experiment. Let the last experiment be a null experiment, i.e. no interventions. This will subject each pair of variables in $G$ to an adjacency test. Hence each pair of variables in $G$ has been subject to at least one directional and one adjacency experiment. Hence, together with the null-experiment, it follows that the causal graph can be completely determined in $\log_2(N) + 1$ experiments. Again, in the case where $N \neq 2^m$, we are dealing with the ceiling of this quantity. □

The worst case always occurs when we have a complete graph among the $N$ variables and we happen - due to bad luck - to intervene upon the vertices in the order of decreasing in-degree, i.e. each randomized vertex is a sink for all vertices not randomized so far. If the graph were not complete, missing edges would be found relatively quickly and this additional information could be exploited. Intervention on a vertex implies that all incoming edges are broken, hence a large in-degree implies that very little information is obtained, since there may be no edge or there may be an incoming edge. However, an outgoing edge would be immediately identifiable due to the dependency between the variables in all possible conditioning sets.

**Lemma 3.4** $\lceil \log_2(N) \rceil$ *experiments are necessary in the worst case to subject all pairs in a causal graph among $N$ variables to a directional test.*

**Proof:** It can easily be shown that $N = 2, 3, 4$ the number of experiments necessary to determine the causal structure is 1, 2 and 2 respectively, i.e. satisfying the above bound.

Suppose the theorem holds for all $N \leq r$. Then let $N = r + 1$. Consider all possibilities for the first experiment $E_1$. It can consist of an intervention on $K$ variables, where $0 \leq K < N = r + 1$. This implies that $E_1$ must subject $K(N - K)$ pairs of variables to a directional experiment. If the underlying true graph is complete, $E_1$ results in a complete undirected graph among the $(N - K)$ variables that were not subject to an intervention and is a zero information experiment for all $\binom{K}{2}$ pairs of variables in the intervened set **I**. Note that $|\mathbf{I}| = K < N = r + 1$ and $|\mathbf{U}| = N - K < N = r + 1$. Hence, we know by the inductive hypothesis that $\log_2(\max(K, N-K))$ experiments are necessary to resolve the remaining subgraphs among variables in **U** and among variables in **V**. So, counting $E_1$, it follows that the total number of experiments necessary to subject all pairs in a causal graph among $N$ variables to a directional test is given by: $x_{total} = 1 + \lceil \log_2(\max(K, N - K)) \rceil = 1 + \lceil \log_2(N/2) \rceil = 1 + \lceil \log_2(N) \rceil - 1 = \lceil \log_2(N) \rceil$. If $N$ is odd, we intervene on $K = \frac{N-1}{2}$, since this maximizes the number of directional tests and also subjects $\binom{(N-K)}{2}$ pairs to adjacency tests, but the above result still holds, since if $N$ is odd, $\lceil \log_2(N) \rceil = \lceil \log_2(N + 1) \rceil$. □

**Theorem 3.5** $\log_2(N) + 1$ *experiments are in the worst case necessary to determine the causal graph among $N$ variables.*

**Proof:** Outline: From Lemma 3.4 it follows that $\log_2(N)$ experiments are necessary to subject each pair of variables to a directional test. The additional experiment results from the fact that not all adjacency tests can be accommodated in $\log_2(N)$ experiments if $N$ is a power of 2. The argument requires some combinatorics, but it should be obvious that for $N = 2$, $\log_2(2) + 1 = 2$ experiments are necessary in the worst case. Lemma 3.4 indicates that in order to achieve all the directional tests within $\log_2(N)$ experiments, one has to intervene on half of the previous intervention set and half of the last passively observed set. The problem for $N = 2^m$ is that this procedure entails that one has to intervene on one variable, say $W$, in every one of the $\log_2(N)$ experiments in the sequence to subject all pairs to one directional experiment. Consequently all pairs of variables $(W, V)$, for some variable $V \neq W$, are only subject to a directional experiment (namely when $V$ is not in the intervention set) and to a zero information experiment (when $W$ and $V$ are in the intervention set). However, this combination of experiments is in the worst case insufficient to determine the edge between the variables (one cannot distinguish between $V \rightarrow W$ and no edge). This problem arises only in the case when $N$ is a power of 2, and in these cases one more (passive observational) experiment is required in addition to the $\log_2(N)$ experiments. When $N \neq 2^m$ the bound on the number of experiments is $\lceil \log_2(N) \rceil$. The last passive observational experiment is not necessary. For any $N \neq 2^m$, given $\log_2(N)$ experiments it is possible to subject all pairs to a directional test and still ensure that there is no variable that is intervened upon in every experiment (see figure 2). This, together with some combinatorics implies the second test for each pair (adjacency or opposing directional) is already accounted for within the initial $\lceil \log_2(N) \rceil$ number of experiments. □

Combining the above results then, we have that $\lfloor \log_2(N) \rfloor + 1$ is a tight bound on the number of experiments for any $N \geq 2$. [1]

**Lemma 3.6** *In order to determine a causal graph of $N$ variables where the number of simultaneous experiments in any one experiment is limited by $k_{max} < \frac{N}{2}$, $(\frac{N}{k_{max}} - 1) + \frac{N}{2k_{max}} \log_2(k_{max})$ are sufficient.*

**Proof:** (Outline) Suppose for the sake of argument that $k_{max}$ divides $N$ by some integer $p$ where $p$ is an even number. [2] Divide the $N$ variables into $p$ disjoint subsets of $k_{max}$ variables. Let the first $p - 1$ experiments each be a $k_{max}$-interventions on one of the $p$ sets of variables. These $p - 1$ experiments will result

---
[1]Since $\lfloor \log_2(N) \rfloor + 1 = \lceil \log_2(N) \rceil$
[2]If not, then we do not have integer values for the bound, however the results still hold for the ceiling of the resulting value.

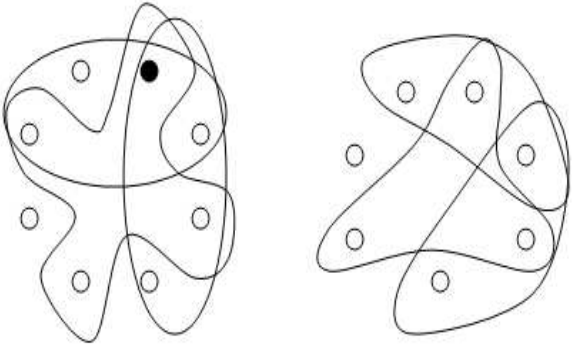

Figure 2: The intervention sets required to subject all pairs of variables to a directional test within $\log_2(N) = 3$ experiments for 8 and 7 variables respectively. While in the case of $N = 8$, there is one variable that is subject to an intervention in every experiment (shown in black), this can be avoided for $N = 7$

in directional tests for all pairs of variables that go between the $p$ sets and in adjacency tests for all pairs of variables in the graph. So after $p-1$ experiments every pair of variables has been subject to one adjacency test and only the pairs of variables within each of the $p$ sets have not yet been subjected to a directional test. From Lemma 3.2 it follows that $\log_2(k_{max})$ experiments are sufficient to subject all pairs in a causal graph among $k_{max}$ variables to a directional test. [3] Since the maximum size of the intervention set used in Lemma 3.2 is (in our case here) $\frac{k_{max}}{2}$ and since we are restricted in this case by $k_{max}$ as the maximum size of the intervention set, it follows that we can perform interventions on two of the $p$ sets concurrently in one experiment. Consequently, $\frac{p}{2} = \frac{N}{2k_{max}}$ sequences of $\log_2(k_{max})$ experiments each are sufficient to subject all the pairs of variables in the $p$ sets to a directional test. Once all these experiments have been performed, every pair of variables in the $N$-graph has been subject to an adjacency and a directional test, which is sufficient to determine the causal graph. □

**Theorem 3.7** *In order to determine a causal graph of $N$ variables where the number of simultaneous experiments in any one experiment is limited by $k_{max} < \frac{N}{2}$, $(\frac{N}{k_{max}} - 1) + \frac{N}{2k_{max}} \log_2(k_{max})$ are sufficient and in the worst case necessary.*

**Proof:** Outline: It follows from Lemma 3.6 that the specified number of experiments is sufficient. Now consider a worst case graph: a complete graph among the $N$ variables where $k_{max}$ divides $N$ by some in-teger $p$ and $p$ is an even number. Since $k_{max} < \frac{N}{2}$ the main concern is to accommodate all the $\binom{N}{2}$ directional tests in as few experiments as possible. Clearly, if the $N$ variables are split into $p$ arbitrary disjoint subsets, then the first $p-1$ experiments, each a $k_{max}$-intervention on a different one of the $p$ sets will subject the maximum number of pairs of variables to directional tests, namely for each experiment, $k_{max}(N - k_{max})$ pairs while no pair is repeated in these $p-1$ experiments. Firstly, note that in the worst case we assume that by lack of luck the interventions occur on the variables in the order of decreasing in-degree, i.e. every intervention is on the common effect of all so far not intervened upon variables, i.e. on the sink. Consequently it is necessary to subject each pair of variables to two tests: one adjacency and one directional test or two opposing directional tests. Further note that while this sequence of $p-1$ experiments subjected the - for this number of experiments - maximum number of pairs of variables to a directional test and thereby subjected all pairs between the $p$ sets to directional tests, it also subjected all pairs of variables in the graph to an adjacency test for free. Hence we are at this point in the worst case scenario only left with complete undirected subgraphs among the $k_{max}$ variables in each of the $p$ subsets we divided the graph into. It should now be clear that it follows from Lemma 3.4 that $\log_2(k_{max})$ experiments are in the worst case necessary to subject all the pairs of variables in a graph of size $k_{max}$ to a directional test. [4] So an additional $\log_2(k_{max})$ experiments are necessary to subject all the remaining pairs to directional tests. However, we cannot run the directional tests on all the $p$ subsets simultaneously since that would require an intervention set of size $\frac{N}{2}$. Hence, we can only consider two subgraphs concurrently, totaling an intervention set size of $k_{max}$. Hence $\frac{p}{2} = \frac{N}{2k_{max}}$ such sequences of experiments are necessary, since we will deal with two of the subgraphs at a time. Hence, in total we require $p - 1 + \frac{p}{2} \log_2(k_{max})$ experiments to determine the causal graph, giving the above result. □

## 4 DISCUSSION

Our results have a variety of limitations. They do not apply when there is prior knowledge restricting the class of possible DAGs. Depending on the prior knowledge, lower bounds are possible. Our results are idealized in the sense that we assume the data perfectly reflect the conditional independence relations implied by the Markov properties of a DAG describing the causal relations. This means the bounds do not apply in con-texts where, for example, the sample, no matter how large, presents a mixture of systems represented by two or more distinct DAGs. That is equivalent to a latent variable problem, discussed below, in which the latent variable values code for different DAGs or probability distributions. Further, the bounds we find are combinatorial rather than statistical, and statistical issues are critical in practice. There are trade-offs that are difficult to assess in any particular context, notably the increased probability of accurate inferences to Markov properties with increasing sample size versus the cost of increasing sample sizes. There are also trade-offs that are difficult to assess in general, such as the relative costs of more experiments with fewer simultaneous randomizations as against fewer experiments with more simultaneous randomizations. Our results are therefore properly understood as the lowest of worst case bounds, against which the efficiency of heuristic procedures or practical strategies might be measured.

---

[3] Note that Lemma 3.2 has no restriction preventing it from applying to a subgraph.

[4] Note that Lemma 3.4 has no restriction preventing it from applying to a subgraph.

As with Murphy and with Tong and Koller, our results do not extend to cases with latent variables, sample selection bias, or feedback. The detailed reasons vary, but the general problem is the same: in such cases, existing search procedures for features of the Markov equivalence class of directed graphs do not in general determine adjacency relations. For systems whose conditional independence relations are represented by the d-separation properties of a cyclic graph, graphs with distinct adjacencies may belong to the same d-separation equivalence class (Richardson, 1996). Pointwise consistent search procedures when latent variables and/or sample selection bias may be present, e.g., the FCI algorithm (Spirtes, et al., 1993; Spirtes, Meek and Richardson, 1999) exploit only conditional independence relations among observed variables, and determine only a partial ancestral graph (Spirtes, Meek and Richardson 1999) or a mixed ancestral graph (Richardson and Spirtes, 2002). So, for example, the partial ancestral graph of the DAG on the left in figure 3, with $V_1, V_2$ and $V_3$ observed, $U$ unobserved, is given on the right in figure 3.

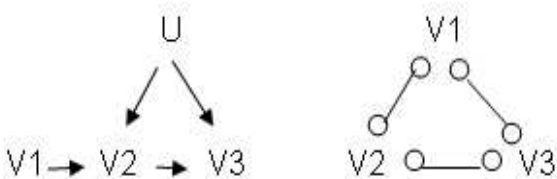

Figure 3: The true graph with $V_1, V_2$ and $V_3$ observed and $U$ unobserved on the left, and its partial ancestral graph on the right.

These complexities do not imply that no characterization of lowest worst case bounds is possible for such problems. For example, suppose a first experiment for data from figure 3 (left) randomizes $V_1$, yielding the result that $V_1$ causes $V_2$ and $V_3$, and $V_2$ and $V_3$ are adjacent, but does not determine whether there is a directed edge from $V_1$ to $V_2$ or from $V_1$ to $V_3$, or both. Then randomizing $V_2$ determines the $V_2 \to V_3$ edge direction, and, because $V_1$ and $V_3$ are independent when $V_2$ is randomized, determines that there is no $V_1 \to V_3$ edge. If, however, in the second experiment, $V_3$ had been randomized instead of $V_2$, only the direction of the $V_2 \to V_3$ edge would be determined, and a third experiment, randomizing $V_2$, would be required. We conjecture that $N$ single experiments always suffice to determine the causal relations among observed variables that are a subset of the vertices of a DAG, and $2 \log_2(N)$ multi-intervention experiments suffice.

**Acknowledgements**

The second author is supported by NASA grant NCC2-1227 and a grant from the Office of Naval Research to the Florida Institute for Human and Machine Cognition for Human Systems Technology. The third author is supported by a grant from the James S. McDonnell Foundation.

**References**

D. M. Chickering (2002). *Learning Equivalence Classes of Bayesian-Network Structures*. Journal of Machine Learning Research, 2:445-498.

F. Eberhardt, R. Scheines, and C. Glymour (2004). *N-1 Experiments Suffice to Determine the Causal Relations Among N Vairables*. In Department of Philosophy, Carnegie Mellon University Technical Report CMU-PHIL-161.

C. Meek, (1996). Ph.D Thesis, Department of Philosophy, Carnegie Mellon University

K. P. Murphy, (2001). *Active Learning of Causal Bayes Net Structure*, Technical Report, Department of Computer Science, U.C. Berkeley.

J. Pearl, (1988). *Probabilistic Reasoning in Intelligent Systems*, San Mateo, CA. Morgan Kaufmann.

T. Richardson, (1996). Ph.D Thesis, Department of Philosophy, Carnegie Mellon University

T. Richardson, and P. Spirtes, (2002). *Ancestral Graph Markov Models*. Annals of Statistics. 30, 4, 962- 1030.

P. Spirtes, C. Meek, and T. Richardson, (1999). *An Algorithm for Causal Inference in the Presence of Latent Variables and Selection Bias* in Computation, Causation and Discovery, 1999; MIT Press.

P. Spirtes, C. Glymour, and R. Scheines, (2000). *Causation, Prediction, and Search*, 2nd ed. New York, N.Y.: MIT Press.

S. Tong, and D. Koller, (2001). *Active Learning for Structure in Bayesian Networks*, Proceedings of the International Joint Conference on Artificial Intelligence.